\documentclass[conference]{IEEEtran}

\newcommand{\publicationnotice}{This paper has been published in the Proceedings of the IEEE International Conference on Machine Learning and Applications (ICMLA 2026). DOI: 10.1109/ICMLA69890.2026.00053.}

\newcommand{\copyrightnotice}{\copyright\ 2026 IEEE. Personal use of this material is permitted. Permission from IEEE must be obtained for all other uses, in any current or future media, including reprinting/republishing this material for advertising or promotional purposes, creating new collective works, for resale or redistribution to servers or lists, or reuse of any copyrighted component of this work in other works.}

\IEEEoverridecommandlockouts
\usepackage{cite}
\usepackage{amsmath,amssymb,amsfonts}
\usepackage{algorithmic}
\usepackage{graphicx}
\usepackage{textcomp}
\usepackage{xcolor}
\usepackage{booktabs}
\usepackage{multirow}
\usepackage[T1]{fontenc}
\def\BibTeX{{\rm B\kern-.05em{\sc i\kern-.025em b}\kern-.08em
    T\kern-.1667em\lower.7ex\hbox{E}\kern-.125emX}}
\begin{document}
\title{Automated Annotation-Free Synthetic Data Generation for Underwater Litter Detection: A Comparison of Generative Approaches\\
\thanks{This publication has been supported by the funding from the European Union's Horizon Europe research and innovation programme under grant agreement No 101093822 (SeaClear2.0).}
}

\author{\IEEEauthorblockN{Amala Mary Vincent}
\IEEEauthorblockA{\textit{Delft Center for Systems and Control} \\
\textit{Delft University of Technology}\\
Delft, Netherlands \\
a.m.vincent@tudelft.nl}
\and
\IEEEauthorblockN{Bart De Schutter}
\IEEEauthorblockA{\textit{Delft Center for Systems and Control} \\
\textit{Delft University of Technology}\\
Delft, Netherlands \\
b.deschutter@tudelft.nl}
}

\maketitle

\begin{center}
\begin{minipage}{0.95\columnwidth}
\footnotesize
\publicationnotice

\medskip
\copyrightnotice
\end{minipage}
\end{center}

\begin{abstract}
Underwater object detection is often limited by scarce and imbalanced training data and by strong appearance variations caused by turbidity, lighting, and differences between sensors. Existing data augmentation strategies, such as geometric or color augmentation or fully synthetic image generation with Generative Adversarial Networks (GANs), either lack semantic realism, fail to respect scene physics, or require additional annotations. In low-data domains like underwater environmental monitoring, the need for additional annotations is often the core bottleneck. Collecting new footage is hard, but labeling this dataset is even more time-consuming and costly, which limits how quickly a usable dataset can be constructed. We propose a unified fully automated zero-annotation pipeline that inserts class-conditioned trash objects into real underwater frames while preserving the original background context. Starting from sampled bounding boxes, each box is converted into an insertion mask and the object is pasted or generated within it, while the same box coordinates are automatically written as the corresponding annotation. Because labels are derived directly from the sampled boxes, the pipeline generates fully labeled training data for low-data settings without any manual annotation. Within this single pipeline we compare three strategies: copy-paste, Stable Diffusion inpainting, and DALL-E image-conditioned generation. We evaluate bounding-box faithfulness, visual realism, and the impact of the generated samples on downstream detector performance on the SeaClear underwater debris dataset. Results show that DALL-E improves detection performance in four of five domains, with relative gains of up to 96 percent over the real-only baseline.
\end{abstract}


\section{Introduction}
\label{sec:intro}

The accumulation of underwater litter poses a critical threat to marine ecosystems, impacting biodiversity and water quality \cite{thiagarajan2025urgent}. Efficient detection and removal of this debris are essential for environmental preservation. However, the underwater environment presents unique challenges for autonomous systems \cite{zhong2025exploring}. Reliable detection typically requires robust computer vision models, yet the development of these models is currently hindered by a significant lack of high-quality, diverse datasets \cite{muzammul2025comprehensive}.

A key challenge is the limited availability of labeled underwater data. Unlike terrestrial datasets, underwater imagery is difficult to capture due to light attenuation, turbidity, and the sheer vastness of the ocean floor \cite{yuan2022survey, naveen2025advancements}. Consequently, there is a severe shortage of annotated images necessary to train deep learning models effectively for underwater object detection, leading to poor generalization in real-world scenarios \cite{er2023research, elmezain2025advancing}.

Simply collecting more raw data does not resolve this bottleneck. Gathering extensive imagery for every unique location is both expensive and time-consuming, making it impractical for large-scale deployment. More critically, even when footage is available, it remains unusable for training until it is annotated: manually identifying and drawing boundaries around degraded, often partially buried or occluded litter demands substantial expert effort. In low-data underwater domains, it is therefore annotation cost, rather than data collection, that fundamentally limits how quickly detection models can be trained and deployed. This is the bottleneck our work directly targets.

To address this issue, we propose a unified fully automated zero-annotation pipeline for generating labeled underwater training data. The pipeline places trash objects at sampled locations within real underwater backgrounds. Each sampled bounding box serves a dual purpose: it acts as the spatial constraint that determines where the object is inserted, and its coordinates are written directly as the corresponding annotation. Because the label is derived from the box itself, no manual labeling is required at any stage of the process. Within this single pipeline, we instantiate and systematically compare three object-insertion strategies: copy-paste of real object crops, Stable Diffusion inpainting, and DALL-E image-conditioned generation. While copy-paste provides guaranteed visual fidelity but limited diversity, SD inpainting offers controlled local editing and DALL-E represents unconstrained scene-aware generation. Therefore this choice enables fair comparison of the effect of synthetic object realism on detection pipeline, with fully automated annotation for all three methods.

The motivation for this work lies in overcoming domain shifts and low-data regimes. Standard augmentation techniques such as rotation or color jittering provide only marginal gains, as they do not introduce new object variations \cite{kumar2024image}. Fully synthetic generation with GANs often struggles with structural consistency in complex underwater scenes \cite{kishan2025underwater}, and still requires the synthetic data to be annotated. Object-insertion approaches avoid the annotation problem by construction, but they differ widely in how realistically the inserted objects blend with the underwater scene. Rather than assuming any one generator is best, we treat realism as an empirical question and measure which insertion strategy actually translates into improved detection under domain shift.

The main contributions of this work are as follows:
\begin{itemize}
\item We propose a fully automated, zero-annotation pipeline in which sampled bounding boxes act simultaneously as spatial generation constraints and as ready-to-use annotations, eliminating manual labeling entirely.
\item We perform a systematic comparison of three object-insertion strategies: copy-paste, Stable Diffusion inpainting, and DALL-E image-conditioned generation, within a single unified pipeline, evaluated across five domains of the SeaClear underwater debris dataset \cite{djuravs2024dataset}. 
\end{itemize}

The remainder of this paper is organized as follows: Section \ref{sec2} presents the related work and Section \ref{sec3} describes the dataset used. Section \ref{sec4} details the synthetic data generation pipeline. Section \ref{sec5} outlines our experimental setup and the evaluation criteria used. Section \ref{sec6} presents the results and a discussion of the observed performance gains. Finally, Section \ref{sec7} concludes the paper and suggests avenues for future research.

\section{Related Work}
\label{sec2}
Early underwater litter detection relied on manual diver surveys and towed-camera
review, which are primarily practical in shallow or accessible environments. The shift to deep learning improved both coverage and throughput: Valdenegro-Toro trained deep networks on forward-looking sonar images to detect submerged debris \cite{valdenegro2019deep}, and Politikos et al. used a Mask R-CNN detector on towed-camera imagery to reach 62\% mAP across eleven litter categories \cite{politikos2021automatic}. These results demonstrate the suitability of deep detectors as the standard tool for underwater litter monitoring, but all of them are trained on manually annotated data, which is exactly the bottleneck this study targets.

To reduce annotation cost, object-insertion augmentation pastes or renders objects into existing background images and derives labels directly from the placement coordinates. Dwibedi et al. showed that even naive "cut-and-paste" of object crops onto random backgrounds gives useful training signal for instance detectors \cite{dwibedi2017cut}, and Ghiasi et al. later showed that large-scale jittering plus random copy-paste is a strong, simple augmentation for instance segmentation on natural images \cite{ghiasi2021simple}. These methods are designed for terrestrial, clutter-controlled datasets (e.g., COCO) and have not been adapted to the low-visibility, low-data underwater debris setting, where background turbidity and small object scale change the trade-offs substantially.

Generative augmentation offers an alternative to copy-paste by synthesizing the inserted object rather than reusing a fixed crop. Latent diffusion models such as
Stable Diffusion support prompt- or mask-conditioned inpainting and can be fine-tuned on domain-specific images \cite{rombach2022high}, while proprietary image-conditioned models such as DALL-E can regenerate scene regions without any fine-tuning \cite{openai_dalle3}. 

To the best of our knowledge, no prior study has directly compared copy-paste augmentation, diffusion inpainting, and image-conditioned generation within a single automated annotation pipeline for underwater debris detection. Previous work has also not established whether distribution-level realism metrics, such as FID and KID, predict downstream detection improvements in this setting. We address this gap by deriving labels directly from the same sampled bounding boxes used for object insertion. This keeps the annotation procedure consistent across all three strategies and helps isolate the effect of object-generation quality on real-world detection performance under domain shift.

\section{Data}
\label{sec3}
\subsection{SeaClear Marine Debris Dataset}
We use the SeaClear Marine Debris Dataset, an underwater dataset annotated for object detection and instance segmentation. The dataset consists of 8610 images captured using various Remotely Operated Vehicles (ROVs) and multiple camera sensors across five geographic locations in Croatia and France, resulting in 11 domains total when accounting for camera and temporal variations \cite{djuravs2024dataset}. Data were collected across five sites reflecting different human activities and underwater conditions: Jakljan, Croatia; Lokrum (Portoč), Croatia; Bistrina (Mali Ston Bay), Croatia; Slano, Croatia; and Marseille (L'Estaque), France.

\subsection{Targeted Low-Data Domains}
To evaluate the synthetic pipeline across diverse underwater regimes and data scarcity levels, we select five representative domains spanning very-low-data to low-data regimes. These five domains are characterized by significant domain shifts in turbidity, debris types, object scales, and environmental conditions:
\begin{itemize}
\item Jakljan II (J2): 65 images, very-low-data regime with small objects. A Croatian islet location with lower debris density, dominated by bottles and cans linked to nautical tourism.
\item Slano II (S2): 71 images, very-low-data regime with small objects. A Croatian harbour village characterized by diverse debris often partially buried in construction rubble, making objects difficult to separate from background.
\item Lokrum II (L2): 77 images, very-low-data regime with small objects. A relatively clear Croatian site with Posidonia oceanica seagrass where litter mainly stems from tourism and nautical activity.
\item Slano I (S1): 168 images, low-data regime with medium-sized objects. The same harbour village as S2 but representing a different section with more medium-sized debris.
\item Bistrina III (B3): 193 images, low-data regime with larger objects and high turbidity. A Croatian site strongly influenced by oyster farming with sediment-rich, highly turbid water and common debris such as plastic trays and nets.
\end{itemize}

\subsection{Annotations}
SeaClear annotations include four semantic classes: Trash (human-made debris: plastic, metal, rubber, glass, fibre), Plant (marine vegetation), Animal (marine fauna), and ROV (underwater robot parts visible in-frame). Since the goal of this work is autonomous litter removal, we focus only on the Trash class. All training, generation, and evaluation steps use the Trash class, and the other classes are not used.

\section{Synthetic Data Generation Pipeline}
\label{sec4}
This section details the fully automated zero-annotation synthetic data generation pipeline that compares three object-insertion strategies to address the scarcity of annotated underwater imagery.

\subsection{Automated Bounding Box Generation and Constraints}

To ensure the generated objects are both realistic and useful for training, we implement a stochastic bounding box generation strategy. To ensure that boxes are large enough for reliable insertion or inpainting and meaningful object placement, we enforce minimum size constraints. When multiple boxes are generated in the same image, we also control overlap between boxes using an IoU threshold. If the overlap between any pair of boxes exceeds the threshold, the box is rejected and regenerated randomly. Boxes that fail the size or overlap constraints are resampled until a valid box is found. If a valid configuration is not found within a fixed number of iterations, the image is skipped to maintain high data quality.

\subsection{Trash-type Selection and Prompt Mapping}

We define nine trash categories based on the most commonly occurring debris types in the SeaClear dataset composition: rubber tire, glass bottle, fiber rope, concrete pipe, plastic bottle, metal can, plastic bag, plastic pipe, and metal wreckage. Each category is mapped to a class-specific text prompt to condition the generative model. For each sampled box, a trash type is selected, and the corresponding prompt token is used to condition the Stable Diffusion and DALL-E models. The copy-paste strategy does not use prompts or trash-type conditioning; it samples a real object crop at random from a pooled crop library of real trash objects extracted from SeaClear and pastes it into the sampled box.

\subsection{Diffusion-Based Inpainting Pipeline}
We instantiate three object-insertion strategies within the same pipeline. Each strategy is run separately on the same source image. They sample the bounding boxes independently and produce automatic labels derived directly from their sampled boxes. 

\begin{itemize}
\item Copy-Paste:
Real trash crops are extracted from annotated SeaClear images to form a crop library. For each sampled box, a crop of the selected trash type is resized to the box and pasted onto the real background. This provides high object realism but limited diversity.

\item Stable Diffusion (SD) Inpainting:
For each sampled and validated box, a binary mask is generated and passed, together with the background image and a class-specific prompt, to a Stable Diffusion inpainting model \cite{rombach2022high}. The model performs latent-space inpainting, blending the generated object into the local lighting, texture, and turbidity. To better match the underwater domain, we additionally fine-tune the inpainting model on 12968 real trash crops extracted from SeaClear, using domain-specific fine-tuning (5 epochs, learning rate 5 x $10^{-6}$, cosine schedule, 8-bit Adam). During inference, inpainting is performed with strength=0.8 (mask influence), guidance scale=7.5 (classifier-free guidance strength), and num inference steps=50 (diffusion steps).

\item DALL-E image-conditioned generation:
A real underwater image and a class-specific prompt are submitted to DALL-E's image generation API \cite{openai_dalle3}. DALL-E regenerates portions of the scene contextually to integrate the trash, producing a complete synthetic image. This strategy produces realistic underwater objects without any model fine-tuning.
\end{itemize}

\subsection{Output Generation and Annotation}
The pipeline produces two outputs per processed image, regardless of insertion strategy: the synthetic image containing the newly inserted trash, and a corresponding annotation file. The annotation coordinates come directly from the sampled boxes, so no manual labeling is required. All inserted objects are mapped to a single Trash class, consistent with the litter-removal objective.

\subsection{Quality Gating and Validation}
To ensure dataset integrity, generated samples are passed through an automated confidence-based quality filter. Each synthetic image is scored by the real-data baseline detector, and images in which the inserted object is not confidently detected (confidence below 0.5) are discarded, since these typically correspond to failed or unrealistic generations. This replaces subjective manual inspection with a reproducible, quantitative criterion. The filter removed between 11\% and 25\% of generated images depending on the domain, with higher removal rates in the harder, more turbid domains.

\section{Experiments and Evaluation Criteria}
\label{sec5}
This section describes how we evaluate the synthetic data generation pipeline comparing three object-insertion strategies in low-data underwater domains. The evaluation is designed to answer two key questions:
\begin{enumerate}
    \item Do the generated samples look plausible and do they respect the intended box placement?
    \item Do the generated samples improve real-world detection performance under domain shift?
\end{enumerate}

\subsection{Experimental Design}

We evaluate the pipeline on five SeaClear domains: J2, S2, L2 (very-low-data, small objects) and B3, S1 (low-data, larger objects). For each domain we compare seven training conditions: a real-only baseline, and the three insertion strategies (copy-paste, SD, DALL-E) each in two variants: synthetic-only, and synthetic combined with a 25\% subset of real-world baseline images. The 25\% real-data subset is used because mixing a small proportion of real data with synthetic data generalizes better than using the full real set, since a large real set can outweigh the smaller synthetic set and limit its contribution during training \cite{sanchez2023experimental}. The test set for each domain consists exclusively of real, held-out images that are never used for training, validation, or synthetic generation, ensuring an unbiased estimate of real-world performance. Since box sampling and generation are stochastic, each generative configuration is run with three random seeds and results are averaged across them. All detectors use the latest  You Only Look Once (YOLO) model - YOLO26n \cite{redmon2016you, ultralytics2026yolo26docs} (nano variant) trained for 100 epochs and all evaluation is performed on the Trash class only.

\subsection{Box Faithfulness}
A critical requirement for synthetic trash generation is that the generative model accurately respects the spatial constraints of the provided bounding boxes. Since our boxes are stochastically sampled, we evaluate `box faithfulness', which represents the degree to which the model produces a non-trivial insertion strictly inside the intended region, using a relative change score.

The bounding box faithfulness $F$ is measured by comparing the generated image with its corresponding real image to estimate how much the model changed the scene. For each sampled bounding box, we compute the average pixel change inside the box and compare it to the average pixel change outside the boxes (i.e., in the background). If the change is clearly larger inside than outside, the edit is localized to the intended region and the box is considered faithful. This indicates the inpainting respects the spatial constraint and avoids unintended background drift.

More specifically, the bounding box faithfulness is defined as
\begin{equation}
\label{eq1}
F = \frac{\Delta_{\text{in}} - \Delta_{\text{out}}}{\Delta_{\text{out}}},
\end{equation}
where $\Delta_{\text{in}}$ denotes the average pixel change inside the target bounding box and $\Delta_{\text{out}}$ denotes the average pixel change outside all bounding boxes (i.e., in the background).

A value $F$ > 0 indicates that pixel changes are concentrated in the target box rather than scattered into the background ($\Delta_{\text{in}} > \Delta_{\text{out}}$). The quality threshold condition $F > 0$ can be used to filter out unreliable generations where the background changed more than the target region.

Box faithfulness is reported for the generative strategies (SD and DALL-E); for copy-paste the inserted crop fills the box by construction, so the measure is not informative.

\subsection{Visual Realism}
To assess whether synthetic samples resemble real underwater imagery, we compute Fréchet Inception Distance (FID) and Kernel Inception Distance (KID). Both metrics evaluate distributional similarity between real and synthetic images using feature embeddings extracted from a pretrained Inception network.

FID fits multivariate Gaussian distributions to the Inception features of real and synthetic images and computes the Fréchet distance between their means and covariances \cite{heusel2017gans}. Lower FID values indicate that the synthetic image distribution is closer to the real image distribution in feature space.

KID measures the discrepancy between real and synthetic feature distributions using the squared Maximum Mean Discrepancy (MMD), typically with a polynomial kernel \cite{binkowski2018demystifying}. Like FID, lower KID values indicate greater similarity between real and synthetic images. Unlike FID, KID provides an unbiased estimator and is generally considered more reliable when the number of samples is limited \cite{binkowski2018demystifying, chong2020effectively}.

\subsection{Detection Performance on Real Test Images}
Detection performance is evaluated only on a separate test split that is not used during training or validation containing real images, from each target domain to ensure a fair assessment. mAP (mean Average Precision) is computed for the Trash class, using mAP@0.5 and mAP@0.5:0.95. mAP@0.5 uses an IoU threshold of 0.5 and is relatively lenient, while mAP@0.5:0.95 averages AP over IoU thresholds from 0.5 to 0.95 in steps of 0.05, providing a stricter and more localization-sensitive measure. 

\section{Results and Discussion}
\label{sec6}
We first verify that both SD inpainting and DALL-E respect the spatial constraints imposed by sampled boxes. Table \ref{table1} reports the median faithfulness score $F$ (Eq. \ref{eq1}) for each strategy separately, across all five domains. In every domain and for both strategies, all generated images pass the faithfulness check ($F$ > 0), meaning the pixel change inside each target box is consistently larger than the change in the surrounding background. Median $F$ ranges from 19.17 to 43.48 for SD inpainting and from 20.74 to 42.45 for DALL-E, with no consistent ordering between the two strategies across domains. SD is higher in three domains, DALL-E in two, and differences never exceed 3 points. Both generative strategies therefore localize their edits comparably well within the target box, despite DALL-E regenerating substantially larger portions of the surrounding scene, as reflected in its far higher whole-image FID and KID (Table \ref{table2}). This shows that DALL-E's broader scene-level drift does not come at the cost of box-level localization and the automatically generated labels remain spatially valid for both strategies.

\begin{table}[t]
\centering
\caption{Box faithfulness of SD and DALL-E generated images across all five domains, reported separately by strategy. Median relative change score $F$ (Eq. \ref{eq1}). Higher $F$ indicates stronger localization of edits within the target box.}
\label{table1}
\begin{tabular}{lcc}
\toprule
\textbf{Domain} & \textbf{Median $F$ for SD} & \textbf{Median $F$ for DALL-E} \\
\midrule
J2 & 25.37 &  22.21\\
S2 & 41.77 &  42.45\\
L2 & 22.12 &  21.43\\
S1 & 43.48 &  40.89\\
B3 & 19.17 &  20.74\\
\bottomrule
\end{tabular}
\end{table}

Table \ref{table2} reports the corresponding FID and KID scores for each strategy.

\begin{table}[t]
\centering
\caption{Visual realism: FID and KID for all strategies across domains. Lower values indicate closer match to real image distributions.}
\label{table2}
\begin{tabular}{lrrrrrr}
\toprule
\textbf{Domain} 
& \multicolumn{2}{c}{\textbf{Copy-paste}} 
& \multicolumn{2}{c}{\textbf{SD Inpaint}} 
& \multicolumn{2}{c}{\textbf{DALL-E}} \\
\cmidrule(lr){2-3}
\cmidrule(lr){4-5}
\cmidrule(lr){6-7}
& \textbf{FID} & \textbf{KID}
& \textbf{FID} & \textbf{KID}
& \textbf{FID} & \textbf{KID} \\
\midrule
J2 & 62.70 & 0.0186 & 38.58 & 0.0025 & 203.00 & 0.0681 \\
S2 & 38.60 & 0.0144 & 31.15 & 0.0028 & 142.51 & 0.0700 \\
L2 & 88.46 & 0.0410 & 63.63 & 0.0109 & 210.92 & 0.0793 \\
S1 & 55.12 & 0.0305 & 30.33 & 0.0056 & 56.76 & 0.0107 \\
B3 & 50.93 & 0.0215 & 26.67 & 0.0024  & 86.13 & 0.0130 \\
\bottomrule
\end{tabular}
\end{table}

\begin{table}[t]
\centering
\caption{Detection performance (mAP@0.5 and mAP@0.5:0.95) on held-out real test images. Baseline is trained on real data only. Synth-only uses synthetic data without real images. +baseline combines synthetic data with 25\% of the real training set. Results are seed-averaged over three seeds per configuration.}
\label{table3}
\begin{tabular}{llcc}
\toprule
\textbf{Domain} & \textbf{Condition} & \textbf{mAP@0.5} & \textbf{mAP@0.5:0.95} \\
\midrule
J2 & Baseline (real-only)        & 0.0641 & 0.0240 \\
J2 & Copy-paste synth-only       & 0.0000 & 0.0000 \\
J2 & Copy-paste + 25\% baseline  & 0.0050 & 0.0020 \\
J2 & SD inpaint synth-only       & 0.0000 & 0.0000 \\
J2 & SD inpaint + 25\% baseline  & 0.0003 & 0.0002 \\
J2 & DALL-E synth-only           & 0.0540 & 0.0367 \\
J2 & DALL-E + 25\% baseline      & 0.0855 & 0.0434 \\
\midrule
S2 & Baseline (real-only)        & 0.1396 & 0.0794 \\
S2 & Copy-paste synth-only       & 0.0000 & 0.0000 \\
S2 & Copy-paste + 25\% baseline  & 0.0000 & 0.0000 \\
S2 & SD inpaint synth-only       & 0.0000 & 0.0000 \\
S2 & SD inpaint + 25\% baseline  & 0.0000 & 0.0000 \\
S2 & DALL-E synth-only           & 0.1989 & 0.1212 \\
S2 & DALL-E + 25\% baseline      & 0.2738 & 0.1661 \\
\midrule
L2 & Baseline (real-only)        & 0.3620 & 0.1480 \\
L2 & Copy-paste synth-only       & 0.0000 & 0.0000 \\
L2 & Copy-paste + 25\% baseline  & 0.0219 & 0.0065 \\
L2 & SD inpaint synth-only       & 0.0000 & 0.0000 \\
L2 & SD inpaint + 25\% baseline  & 0.0018 & 0.0006 \\
L2 & DALL-E synth-only           & 0.2064 & 0.0765 \\
L2 & DALL-E + 25\% baseline      & 0.5028 & 0.1887 \\
\midrule
S1 & Baseline (real-only)        & 0.3090 & 0.1862 \\
S1 & Copy-paste synth-only       & 0.0000 & 0.0000 \\
S1 & Copy-paste + 25\% baseline  & 0.0050 & 0.0020 \\
S1 & SD inpaint synth-only       & 0.0000 & 0.0000 \\
S1 & SD inpaint + 25\% baseline  & 0.0018 & 0.0007 \\
S1 & DALL-E synth-only           & 0.4333 & 0.2143 \\
S1 & DALL-E + 25\% baseline      & 0.4566 & 0.2463 \\
\midrule
B3 & Baseline (real-only)        & 0.4577 & 0.2179 \\
B3 & Copy-paste synth-only       & 0.0000 & 0.0000 \\
B3 & Copy-paste + 25\% baseline  & 0.0000 & 0.0000 \\
B3 & SD inpaint synth-only       & 0.0000 & 0.0000 \\
B3 & SD inpaint + 25\% baseline  & 0.0000 & 0.0000 \\
B3 & DALL-E synth-only           & 0.2443 & 0.1154 \\
B3 & DALL-E + 25\% baseline      & 0.3343 & 0.1442 \\
\bottomrule
\end{tabular}
\end{table}

\begin{table}[t]
\centering
\caption{Object size distributions (\% of image width) across domains and strategies.}
\label{table4}
\renewcommand{\arraystretch}{1.15}
\begin{tabular}{lllll}
\toprule
\textbf{Domain} & \textbf{Real Objects} & \textbf{Copy-paste} & \textbf{SD Inpaint} & \textbf{DALL-E} \\
\midrule
J2 & 5.0--25.8 & 8.2--44.8 & 11.4--44.6 & 7.7--74.1 \\
S2 & 3.5--88.7 & 8.2--44.8 & 11.4--44.6 & 7.2--93.2 \\
L2 & 1.3--99.9 & 8.2--44.8 & 11.4--44.6 & 4.0--75.8 \\
S1 & 3.2--99.8 & 9.9--44.6 & 9.9--44.6  & 5.9--99.7 \\
B3 & 5.8--99.7 & 9.9--45.0 & 9.9--45.0  & 3.5--83.3 \\
\bottomrule
\end{tabular}
\end{table}

\begin{figure*}[ht]
\centering
\includegraphics[width=\textwidth]{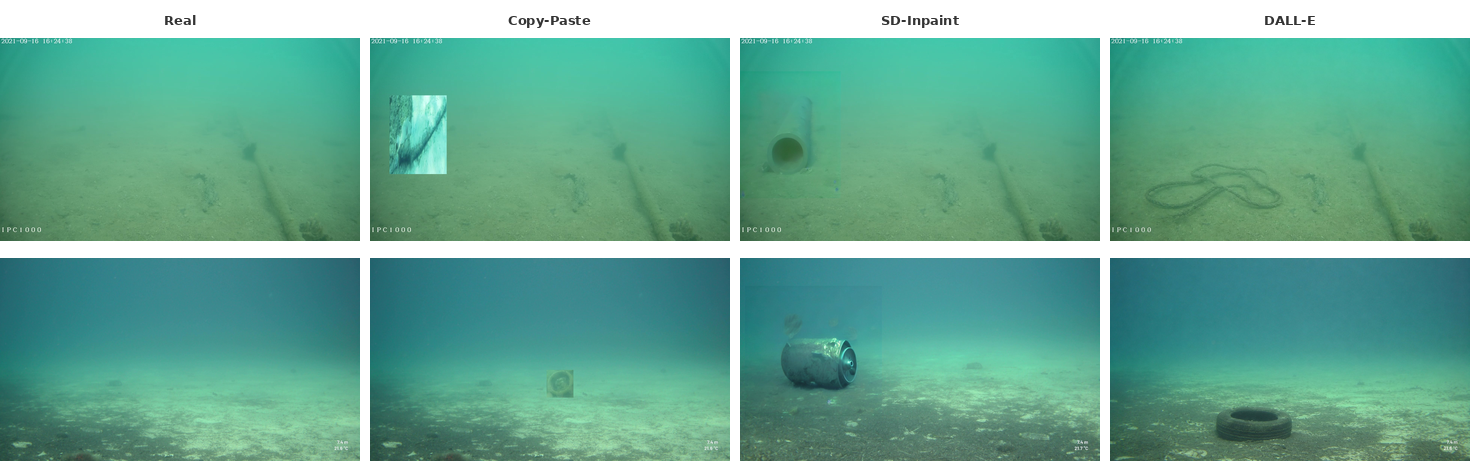}
\caption{Qualitative comparison of object-insertion strategies on B3 and L2 domains.
Two example scenes from different underwater conditions: B3 (larger objects, higher turbidity) in row 1, L2 (small objects, clearer water) in row 2. Each row shows: real image, copy-paste augmentation, SD inpainting, and DALL-E image-conditioned generation.}
\label{fig1}
\end{figure*}

The realism ranking reveals a critical insight: distribution-level realism (FID and KID) does not predict downstream detection performance. The same decoupling appears at the localization level, despite its much higher whole-image FID/KID, DALL-E's edits remain as tightly confined to the sampled box as SD's (Table \ref{table1}), indicating that global distributional drift and local edit precision are independent properties of a generator. SD inpainting achieves the lowest FID and KID across all domains, yet fails to improve detection (see Table \ref{table3}). This occurs because FID and KID measure whole-image statistical similarity, which is dominated by background preservation. SD inpainting preserves the entire real underwater background untouched while inserting objects in a small box, resulting in synthetic images that are statistically very close to real at the whole-image level (low FID and KID), yet the inserted objects are unrealistically large and unrecognizable to the detector. Copy-paste, which also preserves the real background but with sharper discontinuities, scores intermediate. DALL-E image-conditioned generation regenerates portions of the scene, resulting in higher FID and KID values (especially in domains J2 and L2 where FID exceeds 200) because the whole-image statistics drift further from real. However, DALL-E produces object-level realism, i.e., the inserted trash appears plausible with realistic size, texture, and corrosion that is invisible to distribution-level metrics but directly drives detection gains.

The detector does not care about distribution-level statistics; it only cares whether it can recognize the trash object. DALL-E achieves detection gains despite higher FID and KID values. This anti-correlation between FID and KID and detection performance reveals that whole-image metrics are poor proxies for augmentation quality in object insertion settings. Only downstream task performance is a reliable measure of synthetic utility.

SD inpainting and copy-paste fail to improve detection over the baseline in every domain. Even when combined with 25\% real baseline images, both strategies produce near-zero or negligible gains (mAP@0.5 < 0.02 in all cases). This failure is consistent and reproducible across seeds. Analysis of object size distributions (Table \ref{table4}) reveals that SD inpainting and copy-paste have a hard minimum generation size of 8.2–11.4\% of image width, whereas real underwater trash can be as small as 1.3–5.8\%. This prevents generation of the smallest debris objects common in these domains, forcing the detector to learn from a biased, size-skewed synthetic distribution. Additionally, copy-paste suffers from sharp visual discontinuities where crops are pasted, creating unrealistic boundary artifacts. More fundamentally, SD-generated trash items lack realistic underwater appearance, the texture, corrosion patterns, and marine growth that characterize real debris. The failure of copy-paste confirms that constraints on generation, combined with lack of visual realism, are barriers to success. DALL-E, by contrast, can generate objects across the full real size range (minimum 3.5–7.7\%) and with realistic underwater appearance (Table \ref{table4}), enabling successful detection transfer.

DALL-E image-conditioned generation improves detection over the baseline in four of five domains. The one exception is B3, where DALL-E (0.33) falls short of baseline (0.45). B3 is characterized by larger objects and higher turbidity, where the baseline already achieves strong performance; additional gains are limited. However, DALL-E still provides a non-zero contribution when combined with baseline real data, whereas copy-paste and SD inpainting provide none. The key difference is object generation quality. DALL-E produces smaller, more realistically integrated trash objects that respect the size distribution of real underwater debris. Its whole-image FID is high (142–211) because it regenerates portions of the scene for contextual blending, but this global statistical drift is invisible to the downstream detector, which cares only about object-level realism. The realism difference is visible qualitatively in Figure \ref{fig1}.

Overall, synthetic data alone is insufficient across all strategies in most domains. Synth-only training uniformly produces near-zero detection performance (mAP@0.5 < 0.05 except DALL-E on S1 where it reaches 0.43). The detector does not generalize from synthetic-only data to real test imagery without any real training examples. Mixing a small proportion of real data (25\% of the baseline training set) stabilizes learning and enables the detector to ground synthetic patterns in real-world appearance. Even so, the dominant factor determining success is synthetic object realism, not the mix ratio: DALL-E + 25\% baseline vastly outperforms copy-paste + 25\% baseline or SD inpaint + 25\% baseline in every domain. This underscores the finding that object-level realism, not the quantity of real data in the training mix, is the bottleneck.

\section{Conclusions and Future Work}
\label{sec7}
In this study, we have proposed a fully automated, zero-annotation pipeline for generating labeled synthetic training data to address the annotation bottleneck in low-data underwater litter detection. By deriving YOLO labels directly from sampled bounding boxes, the pipeline eliminates manual annotation entirely, which is a critical advantage in data-scarce domains. Within a unified framework, we have systematically compared three object-insertion strategies: copy-paste, Stable Diffusion inpainting, and DALL-E image-conditioned generation, across five underwater domains of SeaClear data.

Our key finding is that object-level realism, not distribution-level metrics, determines detection success. DALL-E image-conditioned generation achieves the only consistent detection improvements (4 of 5 domains, up to +96\% relative mAP@0.5, example, for S2 domain from 0.14 to 0.27) despite scoring worst on FID and KID metrics, because it produces realistic trash objects. Copy-paste and SD inpainting fail uniformly across all domains, not due to label errors, but because they generate low quality trash images that are integrated into the real image and the object size exceeds real object size distributions. This reveals that conventional realism metrics are misleading in trash generation settings, and the appearance of the trash items is the deciding factor.

A limitation of this study is that DALL-E is a closed, proprietary model whose architecture, training data, prompt-processing mechanisms, and generation procedures are not publicly disclosed. This limits the interpretability of why DALL-E outperforms the open-source strategies and may constrain reproducibility if the model, API, or prompt-processing behavior changes.

Directions for future work include:
\begin{itemize}
\item The zero-annotation synthetic augmentation approach can be integrated into a complete semi-autonomous pipeline integrating synthetic bootstrapping with active learning and iterative retraining.

\item The zero-annotation pipeline can be evaluated on diverse tasks beyond litter detection.
 
\end{itemize}

\bibliographystyle{IEEEtran}
\bibliography{refs}

\end{document}